\documentclass[11pt,letterpaper]{article}
\usepackage{emnlp2017}
\usepackage{times}
\usepackage{latexsym}
\usepackage[utf8]{inputenc}
\usepackage{amssymb,amsfonts,amsmath}
\usepackage{graphicx}
\usepackage{multirow}
\usepackage{enumitem}

\usepackage{float}
\usepackage{wrapfig}
\restylefloat{figure}
\usepackage{booktabs}
\usepackage[english]{babel}

\usepackage{caption}
\usepackage{subcaption}
\usepackage{csquotes}

\usepackage{setspace}
\usepackage{array}
\usepackage{calc}
\usepackage{multicol}
\usepackage[subpreambles=true]{standalone}

\usepackage{pgfplots}
\usepackage{pgfkeys}
\usepgfplotslibrary{dateplot} 
\pgfplotsset{compat=newest}
\usetikzlibrary{matrix}
\usepgfplotslibrary{groupplots} 
\usetikzlibrary{patterns}
\usepgfplotslibrary{external} 
\pgfplotsset{
  invoke before crossref tikzpicture={\tikzexternaldisable},
  invoke after crossref tikzpicture={\tikzexternalenable},
}

\usepackage[capitalize]{cleveref}

\emnlpfinalcopy

\crefformat{appendix}{#2#1#3}

\DeclareMathOperator*{\argmin}{arg\,min~}

\newcommand{\x}{\mathbf{x}}
\newcommand{\y}{\mathbf{y}}
\newcommand{\h}{\mathbf{h}}
\newcommand{\VTheta}{\mathbf{\Theta}}

\newcommand{\vm}{\mathbf{m}}
\newcommand{\vv}{\mathbf{v}}

\title{Online Learning for Neural\\ Machine Translation Post-editing}

 \author{\'{A}lvaro Peris \and Luis Cebri\'{a}n \and Francisco Casacuberta \\
         Pattern Recognition and Human Language Technology Research Center \\ Universitat Politècnica de València, València, Spain\\
        \texttt{\{lvapeab, fcn\}@prhlt.upv.es} \\
        \texttt{luicebch@inf.upv.es } }
        
\date{}

\begin{document}

\maketitle

\begin{abstract}
Neural machine translation has meant a revolution of the field. Nevertheless, post-editing the outputs of the system is mandatory for tasks requiring high translation quality. 
Post-editing offers a unique opportunity for improving neural machine translation systems, using online learning techniques and treating the post-edited translations as new, fresh training data.
%We delve into the application of online learning in the neural machine translation post-editing process, aiming to reduce the required effort. 
We review classical learning methods and propose a new optimization algorithm. We thoroughly compare online learning algorithms in a post-editing scenario. Results show significant improvements in translation quality and effort reduction.
\end{abstract}

\section{Introduction} \label{sec:introducion}

In recent years, the statistical machine translation (SMT) field has evolved by leaps and bounds. Nevertheless, SMT systems still require human supervision in order to produce high-quality translations. This correction process is known as post-editing. In the machine translation (MT) industry, the reduction of the post-editing effort has a great interest, as it leads to larger productivity \citep{Arenas08,Plitt10}.

The text obtained during the post-editing process can be converted into new training data, useful for adapting the system to a different domain or to a changing environment.

Online learning (OL) techniques allow the system to learn from post-edited samples and, aiming to avoid the committed mistakes.

The application of OL to the classical phrase-based SMT systems has been thoroughly studied~\citep{Ortiz16}. Nevertheless, over the last years, the new neural machine translation (NMT) technology has shaken the machine translation field~\citep{Sutskever14,Bahdanau15,Wu16}. Therefore, there is a need to reduce the post-editing effort in NMT systems.

The neural approach to MT has a short but meteoric history. The first fully neural translation system was introduced by~\citet{Kalchbrenner13}, being the first competitive NMT systems simultaneously developed by~\citet{Cho14} and \citet{Sutskever14}. From there to now, NMT has surpassed the classical phrase-based systems in many tasks and currently is one the most active research topics (e.g. \citet{Erk16}). Furthermore, the translation industry is also shifting to this new translation paradigm~\citep{Crego16,Wu16}.

NMT relies on the encoder-decoder framework. An encoder recurrent neural network (RNN) processes the source sentence, computing a dense, continuous representation of it. From this representation, a decoder RNN generates, word by word, the translated sentence. In order to cope with the vanishing gradient problem~\citep{Bengio94}, such RNNs feature gated units, such as long short-term memory (LSTM) \citep{Hochreiter97} or gated recurrent units (GRU)~\citep{Cho14}.

The vanilla encoder-decoder model has been steadily improved. The inclusion of attention mechanisms~\citep{Bahdanau15,Luong15a} overcame the problem of processing long source sentences. The management of out-of-vocabulary words has also been explored~\citep{Jean15,Luong15b}. Closely related to this, the NMT vocabulary size limitation has been tackled either developing strategies for taking into account large vocabularies~\citep{Jean15} or shifting from word-level translations to character \citep{Chung16} or sub-word-level \citep{Sennrich16} translations.

Such NMT systems are usually trained from parallel corpora by means of stochastic gradient descent (SGD)~\citep{Robbins51}. SGD can be applied sample by sample, fitting into the OL paradigm.
In this paper, we evaluate the most common SGD optimizers. Moreover, we propose a new SGD variant for NMT, based on passive-aggressive (PA) techniques~\citep{Crammer06}. This method aims to apply the minimum modification to the model (passiveness) required to satisfy a correctness criterion (aggressiveness). The contributions of this work are the following:

\begin{itemize}[noitemsep]
\item We study the application of OL to the NMT framework. To the best of our knowledge, this is the first work that applies OL to NMT in a post-editing scenario.
\item We present a new OL algorithm, inspired by PA techniques and implemented using subgradient methods.

\item We conduct a wide experimentation, testing the NMT engine in three different scenarios. We compare well-established learning algorithms together with the newly proposed PA algorithms.

\item Results show that OL algorithms are able to significantly reduced the post-editing effort required by an NMT system. Moreover, we demonstrate the capability of NMT for adapting to different domains by means of OL techniques.

\item In order to make research reproducible, we make public the source code of the NMT system and the training algorithms\footnote{ \url{https://github.com/lvapeab/nmt-keras/tree/interactive_NMT}}.

\end{itemize}

The rest of the paper is structured as follows: after this introduction,~\cref{sec:related-work} reviews the related work. An NMT system is briefly described in~\cref{sec:nmt}. The classical SGD methods are revisited in~\cref{sec:online-sgd}.~\cref{sec:pa-subgrad} details the newly proposed online algorithm. \cref{sec:exp-setup} presents the experimental framework designed to assess our method. Results are shown and discussed in~\cref{sec:results}. Finally, we conclude in~\cref{sec:conclusions}, tracing future research lines.
\section{Related work} 
\label{sec:related-work}

Online learning is a profusely studied topic in machine learning. In the field of SMT, OL techniques are mainly employed for tuning the weights of the log-linear model~\citep{Zens02,Och02} from classical phrase-based SMT systems~\citep{Koehn10}. More specifically, PA-based techniques, such as the margin infuse relaxed algorithm (MIRA)~\citep{Crammer01} are especially well-suited when dealing with a large number of features. Therefore, MIRA is usually applied to models dealing with lots of sparse features~\citep{Watanabe07,Chiang12}.

Estimating the model features---in addition to the weights of the log-linear model---is a less explored application of OL techniques. Most relating this, were developed under the umbrella of the CasMaCat project~\cite{Alabau13}. As colophon of it,~\citet{Ortiz16} developed an incremental derivation of the EM algorithm. 

Another application of OL techniques in SMT is the adaptation of a trained system to a post-editing scenario. \citet{Martinez-Gomez12} and \citet{Mathur13} used OL for adapting a SMT under a CAT framework. These works are close to ours in the scenarios on which the MT systems are applied. Nevertheless, in our work, we employ an NMT system.

\section{Background: NMT} 
\label{sec:nmt}

In this work, we use an attentional NMT system similar to the one described by~\citet{Bahdanau15}, but using LSTM networks. The input is a sequence of words $\x$ in the source language. Each word is linearly projected to a continuous space by means of an embedding matrix.

The sequence of word embeddings feeds a bidirectional LSTM, which analyzes the input sequence in both directions, from left to right and vice versa. This network computes a sequence of annotations by concatenating the hidden states from the forward and backward layers.

At each decoding timestep, an attention mechanism weights each element from the sequence of annotations, according to the previous decoding state, and computes a joint context vector. 

The decoder is another LSTM network that takes into account the context vector, the previously generated word and its previous hidden state. Finally, a deep output layer~\citep{Pascanu14} is employed to compute a distribution over the target language vocabulary.

 At decoding time, the model approximates the target sentence with a beam-search method~\citep{Sutskever14}.
 
\subsection{Training}

To estimate the model parameters $\VTheta$ (i.e. the weight matrices), the training objective is to minimize a loss function $L_\VTheta$, usually  the minus log-likelihood over a bilingual parallel corpus ${\cal T} = \{(\x_t, \y_t)\}_{t=1}^T$, consisting of $T$ source--target sentence pairs ($\x_t$ and $\y_t$, respectively), with respect to $\VTheta$:
\begin{equation}
\begin{gathered}
\label{eq:training-obj}
\widehat{\VTheta} = \argmin_{\VTheta} L_\VTheta = \\
 \argmin_{\VTheta} \sum_{t=1}^T \sum_{i=1}^{I_t} 
 -\log(p(\y_t \mid {\y_{t, <i},  \x_t}; \VTheta))
\end{gathered}
\end{equation}
where $I_t$ is the length of the $t$-th target sentence.

SGD techniques are the predominant way of optimizing neural networks. The goal is to minimize a given loss function $L_\VTheta$, parametrized by the network parameters. 
%In the case of NMT, the loss is the conditional probability of target sentences given the source sentences (see~\cref{eq:training-obj}).}

SGD updates the parameters in the opposite direction of the gradient of $L_\VTheta$. The update size is controlled by a learning rate. According to when this updates occur, SGD techniques can be classified in batch, minibatch or online modes. Moreover, depending on how the gradient is computed and applied, a large variety of SGD variants have been proposed~\citep{Duchi11,Zeiler12,Kingma14}.
\section{Online gradient descent}
\label{sec:online-sgd}

Online SGD updates the model parameters after each sample. Therefore, for a single sample ($\x_t, \y_t$) the parameters update is computed as:
\begin{equation}
\label{eq:grad}
\Delta \VTheta_t = - \rho~ \nabla L_{\VTheta_t}(\x_t,\y_t)
\end{equation}
where $\rho$ is the learning rate that controls the step size and $\nabla L_{\VTheta_t}$ is the gradient of the objective function $L$ with respect to $\VTheta_t$. Most optimizers used in deep learning, are variants of this vanilla SGD. Therefore, they can also be applied under the OL framework.
In this work, we compared Adagrad, Adadelta and Adam, whose update rules are briefly commented below.

Adagrad~\citep{Duchi11} aims to perform larger updates for infrequent parameters, defining its update rule as:

\begin{displaymath}
\label{eq:grad}
\Delta \VTheta_t = - \frac{\rho}{\sqrt{\mathbf{G}_t + \epsilon}} \odot \nabla L_{\VTheta_t}(\x_t,\y_t)
\end{displaymath}
where $\mathbf{G}_t$ is a vector that contains the sum of the squares of the past gradients with respect to $\VTheta$, $\odot$ is the element-wise multiplication, the division is also applied element-wise and $\epsilon$ is a small value, introduced for numerical stability reasons.

Adadelta~\citep{Zeiler12} is a less aggressive variant of Adadelta, which avoids its constant learning rate decay. Adadelta updates the parameters according to the root mean square (RMS) of the gradients and corrects these updates according to the RMS of the previous update:

\begin{displaymath}
\label{eq:grad}
\Delta \VTheta_t = - \rho \frac{RMS(\Delta \VTheta_{t-1})}{RMS(\nabla L_{\VTheta_t}(\x_t,\y_t))}\nabla L_{\VTheta_t}(\x_t,\y_t)
\end{displaymath}

The default value for $\rho$ is 1. Nevertheless, in an OL scenario, this value must be lowered, in order to perform less aggressive updates (see~\cref{sec:results}).

Finally, Adam~\citep{Kingma14} computes decaying averages for the past gradients ($\vv_t$) and the past squared gradients ($\vm_t$). Such averages are computed following:
$$\vm_t = \beta_1 \vm_{t-1} + (1-\beta_1)\nabla L_{\VTheta_t}(\x_t,\y_t)$$
$$\vv_t = \beta_2 \vv_{t-1} + (1- \beta_2 \nabla L_{\VTheta_t}^2(\x_t,\y_t))$$
where $\beta_1$ and $\beta_2$ are two constant values.
The parameter update performed by Adam is:

\begin{displaymath}
\label{eq:grad}
\Delta \VTheta_t = - \frac{\rho}{\sqrt{\hat{\vv}_t} + \epsilon}\hat{\vm}_t
\end{displaymath}
where $\hat{\vv}_t = \frac{\vv_t}{1-\beta_2^t}$ and $\hat{\vm}_t = \frac{\vm_t}{1-\beta_1^t}$. The divisions and the square root are computed element-wise and $\beta^t$ denotes $\beta$ powered to $t$.

\section{Passive-aggressive via subgradient techniques}
\label{sec:pa-subgrad}
We propose a new version of SGD, inspired by PA techniques such as MIRA~\citep{Crammer01}. The method aims to perform the minimum modification to the model parameters for compelling  with a correctness criterion. Our criterion is directly based on the model loss function (\cref{eq:training-obj}). Nevertheless, we could employ other loss functions, such as BLEU, although this could be costly. We declined towards this loss due to efficiency reasons.

Let %$\x_t$ be a source sentence, 
$\h_t$ be the hypothesis generated by the NMT (using the current parameters $\VTheta_t$) % and $\y_t$ a post-edited target sentence
of the source sentence $\x_t$. We consider that $\VTheta_t$ is incorrect if the model assigns a lower probability to the target reference sentence $\y_t$ than to $\h_t$:
\begin{equation}
p_{\VTheta_t}(\y_t\mid\x_t) < p_{\VTheta_t}(\h_t\mid\x_t)
\end{equation}

In this case, we want to search for a $\widehat\VTheta$ such that $\widehat\VTheta$ is close to $\VTheta_t$ and $p_{\widehat\VTheta}(\y_t\mid\x_t) > p_{\widehat\VTheta}(\h_t\mid\x_t)$. This can be expressed with the loss function $\ell$:
\begin{equation}
\begin{gathered}
\ell(\widehat\VTheta,\x_t,\y_t,\h_t) =  \\
\log p_{\widehat\VTheta}(\h_t\mid\x_t)-\log p_{\widehat\VTheta}(\y_t\mid\x_t) \leq \xi
\end{gathered}
\end{equation}
being $\xi \geq 0$ a slack variable, included for providing more flexibility to the method.

This can be formulated as a minimization problem with constraints:
\begin{equation}
\begin{gathered}
\label{eq:optimizaton-subgrad}
\widehat\VTheta=\argmin_{\VTheta} \frac{1}{2} \|\VTheta-\VTheta_t\|^2+C~\xi \\
	\mbox{s.t.}~~\ell(\VTheta,\x_t,\y_t,\h_t) \leq \xi~~\mbox{and}~~\xi\geq 0
\end{gathered}
\end{equation}
where $C$ is a parameter that controls the aggressiveness of the algorithm \citep{Crammer06}.

From~\cref{eq:optimizaton-subgrad}, $\displaystyle\xi\geq\max(0,\ell(\VTheta,\x_t,\y_t,\h_t))$. We define $F_t$ as the function to optimize:
\begin{displaymath}
\begin{gathered}
F_t(\VTheta,\x_t,\y_t,\h_t) = \\
\frac{1}{2} 
\|\VTheta-\VTheta_t\|^2
+C~\max(0,\ell(\VTheta,\x_t,\y_t,\h_t))
\end{gathered}
\end{displaymath}
aiming to find the set of parameters that minimize this function, i.e.:
\begin{equation}
\widehat\VTheta=\argmin_{\VTheta}F_t(\VTheta,\x_t,\y_t,\h_t)
\end{equation}

For obtaining $\widehat\VTheta$, we use a subgradient method \citep{Shor03}. Subgradient methods are iterative algorithms aimed to solve minimization problems with non-differentiable objective functions. Since our function has discontinuity points, such techniques are well-suited. 
Therefore, we iteratively apply the following weight update:
\begin{equation}
\label{eq:update}
	\Delta \VTheta_t =  -\rho~ \partial_\VTheta F_t(\VTheta,\x_t,\y_t,\h_t)|_{\VTheta^{k}}
\end{equation}
where $\rho$ is the learning rate and $\partial_\VTheta F_t$ is the subgradient of $F_t$ with respect to $\VTheta$. We initialize $\VTheta^{k=0}=\VTheta_t$. This update is applied $k$ times, until reaching some convergence criterion. We denote this passive-aggressive via subgradient methods update rule as PAS.

\subsection{Projected subgradient}
\label{sec:ppa-subgrad}
An extension of the PAS method is the projected subgradient method (PPAS), in which the optimization problem is reformulated~\citep{Boyd03}. Let us define $G_t(\VTheta,\x_t,\y_t,\h_t)$ as $\max(0,\ell(\VTheta,\x_t,\y_t,\h_t))$. Then, \cref{eq:optimizaton-subgrad} can be rewritten as:
\begin{equation}
\begin{gathered}
\widehat\VTheta~=~
\argmin_{\VTheta} G_t(\VTheta,\x_t,\y_t,\h_t)) \\
~~\mbox{s.t.}~~ \|\VTheta-\VTheta_t\|^2 \leq C
\end{gathered}
\end{equation}

With this expression, we can compute the intermediate weight update:
\begin{gather*}
	\bar\VTheta^{k+1}= 
 \VTheta^{k} -
 \rho~ 
 \partial_\VTheta G_t(\VTheta,\x_t,\y_t,\h_t)|_{\VTheta^{k}}
 \end{gather*}
 
 As in the previous case, we initialize $\VTheta^{k=0}=\VTheta_t$ and iterate following some convergence criterion. The final weight update defined as:
\begin{equation}
	\Delta \VTheta_t  =
\frac{\bar\VTheta^{k+1}-\VTheta_t}
{\|\bar\VTheta^{k+1}-\VTheta_t\|}~C
\end{equation}

\section{Experiments and results} 
\label{sec:exp-setup}
In this section, we describe experimental setup used to conduct the experimentation. We define the assessment metrics, describe the corpora and detail the configuration of the NMT systems.

\subsection{Experimental framework}

We tested our OL methods in a post-editing task. Within this scope we define three different scenarios: 1. We are post-editing samples from a different domain to that on which our system has been trained. 2. We are post-editing the outputs of a system trained on the same domain of our task, but we have available out-of-domain data. 3. We only have the in-domain data, either for training and testing.

For each scenario, we adapt the system on-the-fly, applying OL to the post-edited sentences from the test set. The final goal is to reduce the post-editing effort required by upcoming samples. The main difference is that in the first case, we exclusively rely on the OL techniques for performing an adaptation of the system to the translation domain; while in the later cases, the NMT systems are already adapted. We measure up to what extent can OL refine this adapted system. Note that we work under a pure OL framework, i.e., each training sample is seen only once.

In a real post-editing scenario, the system produces a translation hypothesis. The user reviews it and corrects the mistakes committed by the system. To test our proposals in this real post-editing scenario it would require real users, which makes it prohibitively costly. Therefore, we simulate the post-editing process by using the reference sentences as post-edited ones.

In this simulation, the system generates a hypothesis for a given source sentence. Next, this hypothesis is corrected, producing a new training sample (i.e., the reference sentence). The system updates its parameters with this new sample and advances to the next sentence. This procedure is repeated until the translation process is finished. 

\subsection{Metrics and evaluation}

For approximating the post-editing effort, we used the translation edit rate (TER)~\citep{Snover06}. This metric computes the edit distance between hypotheses and reference sentences. The edit operations considered are addition, deletion, substitution and shifting. Moreover, we used BLEU~\citep{Papineni02} and Meteor~\citep{Lavie09} for assessing the translation quality of the systems. BLEU compares the ratio of n-gram structures shared between the system hypotheses and reference sentences, while Meteor computes the F1 score of precision and recall between hypotheses and references, additionally considering exact, stemmed, synonyms and paraphrase matches. For all results shown in this work, we compute $95\%$ confidence intervals by means of bootstrap resampling \citep{Koehn04}.

The evaluation procedure is as follows: for a given source sentence, we produce its translation. This translation is stored for the posterior evaluation. Next, we take the post-edited sentence (i.e. the reference sentence) and update the NMT system with it. This is repeated until the full set of sentences are translated. We evaluate the stored hypotheses, hoping for a steady improvement as the translation process goes on. 

\subsection{Corpora}
\label{sec:corpora}
As out-of-domain data (scenarios 1 and 2), we use the well-known, publicly available Europarl~\citep{Koehn05} corpus, using the partition \texttt{newstest2013} as development set. As in-domain corpora (scenarios 2 and 3), we use the Emea~\citep{Tiedemann09}, XRCE~\citep{Barrachina09} and TED~\citep{Federico11} corpora. The domains of them are medical, printer manuals and TED talks, respectively. We used the standard partitions for all corpora. We used the English--French language pair for all experiments. \cref{tab:corpora} shows the main figures of the corpora.

\begin{table*} [!ht]
	\caption{\label{tab:corpora} Corpora number of sentences ($\vert S \vert $) and vocabulary sizes ($\vert V \vert $). k stands for thousands and M for millions.}
	\centering
	\small
	\begin{tabular}{l lll lll lll}
		\toprule
		 & \multicolumn{3}{c}{Training} &  \multicolumn{3}{c}{Development}  & \multicolumn{3}{c}{Test} \\
		 \cmidrule(lr){2-4} \cmidrule(lr){5-7} \cmidrule(lr){8-10}
 		 & \multirow{2}{*}{$\vert S \vert $} &  \multicolumn{2}{c}{$\vert V \vert $} & \multirow{2}{*}{$\vert S \vert $} &  \multicolumn{2}{c}{$\vert V \vert $} & \multirow{2}{*}{$\vert S \vert $} &  \multicolumn{2}{c}{$\vert V \vert $} \\
		 \cmidrule(lr){3-4}  \cmidrule(lr){6-7}  \cmidrule(lr){9-10}
 		 &  &  \multicolumn{1}{c}{En} &  \multicolumn{1}{c}{Fr} & &  \multicolumn{1}{c}{En} &  \multicolumn{1}{c}{Fr} && \multicolumn{1}{c}{En} &  \multicolumn{1}{c}{Fr}  \\
		\midrule
		Europarl & 2.0M &134k &153k & 3k & 10.5k & 11.9k & -- & -- & -- \\
		Emea & 319k & 52.0k & 59.3k & 500 &2.8k & 2.9k & 1.0k &4.5k & 4.5k\\
		TED & 159k & 46.7k & 58.2k&887 &3.4k & 3.9k &1.7k & 4.9k& 4.9k\\
		XRCE & 52k & 14.0k &15.5k &984 &1.8k&  1.9k& 994 &1.7k &1.8k\\
		\bottomrule
	\end{tabular}
\end{table*}

We tokenized the text using the script from Moses, 
 keeping sentences truecase. For all tasks, we use the joint byte pair encoding (BPE) algorithm for translating at a sub-word level, as described by~\citet{Sennrich16}. We learned 32,000 merge operations.
 In addition to address the unknown word problem, sub-word-level translation allows us to naturally apply an NMT system from one domain to another. We extracted the BPE codes from the out-of-domain corpus (Europarl) and applied them to each in-domain corpus. Therefore, each corpus is segmented according to the out-of-domain corpus. By using this strategy, the vocabulary coverage is extremely high (more than 99.5\% in all cases).

\subsection{NMT systems}

The NMT\footnote{Implementation publicly available at: \url{https://github.com/lvapeab/nmt-keras/tree/interactive_NMT}} systems and OL algorithms were implemented with the Keras and Theano~\citep{Theano16} libraries. 

For choosing the main hyperparameters of the system, we take advantage of the vast exploration made by~\citet{Britz17}. Therefore, as described in~\cref{sec:nmt}, the system consists in an encoder-decoder LSTM network equipped with the attention mechanism described by \citet{Bahdanau15}. For practical reasons, we used single-layered LSTMs. The size of each LSTM is 512, as well as the word embedding size and the attention mechanism layer. As regularization methods, we applied layer normalization~\citep{Ba16} and Gaussian noise to the weights ($\sigma =0.01$)~\citep{Graves11}. We also early stopped the training if the BLEU on the development set did not improve in $100,000$ updates. The original NMT systems were trained using Adadelta~\citep{Zeiler12}, with the default parameters. The norm of the gradients was clipped to 1 in order to avoid the exploding gradient problem~\citep{Pascanu13}. The size of the beam was set to 6 in all experiments.

Online learning hyperparameters were estimated on the Europarl validation set through a grid search, with $\rho$ and $C$ taking the values $10^{-a}, a \in {\{0, -1, -2, -3, -4, -5, -6\}}$. We set $\epsilon = 10^{-8}$ and the rest of the Adam values to their default ($\beta_1=0.9$, $\beta_2=0.999$). \cref{tab:hyperparams} shows the value of the hyperparameters for each algorithm.

\begin{table}
	\caption{\label{tab:hyperparams} Algorithm hyperparameters. $\rho$ refers to the learning rate and $C$ to the aggressiveness of PA algorithms.}
	\centering
	\footnotesize
		{\setlength{\tabcolsep}{4pt}
	\begin{tabular}{lllllll}
		\toprule
		& SGD & Adagrad & Adadelta & Adam & PAS & PPAS \\
		$\rho$ & $10^{-3}$ & $10^{-4}$ & $10^{-1}$ & $10^{-3}$ & $1$  & $10^{-2}$ \\
		$C$ & -- &-- & --& --& $10^{-2}$ &  $10^{-2}$\\
		\bottomrule
	\end{tabular}}
\end{table}

\section{Results and discussion}
\label{sec:results}

In this section, we show and analyze the experimentation conducted. We applied OL to the NMT system with the post-edited samples from the test set, with the goal of reducing the post-editing effort required by the following samples. As stated in \cref{sec:corpora}, we pose three different scenarios:

\begin{enumerate}[noitemsep]
\item We have a general corpus and no in-domain data. Therefore, in order to reduce the post-editing effort, we must take advantage exclusively from the in-domain test samples by means of OL.

\item We have the general corpus, but we also have enough in-domain data for training a system. In this case, we first train with the out-of-domain corpus and next with the in-domain corpus. OL can also fine-tune the resulting system, for obtaining a more tailored system and consequently, reducing the post-editing effort.

\item We only have in-domain corpora. This situation is more artificial than the previous ones, but illustrates up to which extent the NMT system can be improved and adapted to a test set by means of OL.

\end{enumerate}

Since the final scope of this work is a real application, we should check whether the OL algorithms consume an excessive amount of time. The mean update time\footnote{Tested on a Nvidia GTX1080 GPU with CuDNN 5005.} was similar for each algorithm (approximately 65 ms), which is affordable.

\subsection{Adapting without in-domain}

In this scenario, we assume that we only have an out-of-domain corpus. The text to translate comes from an unknown domain. Therefore, we must adapt the system on-the-fly, according to the post-edited sentences: as soon as a sentence is translated and corrected, we modify our system, in order to take this sample into account. As baseline, we take an offline system, which remains static along the translation process.

\begin{figure}
	\centering
	\includegraphics[width=0.5\textwidth]{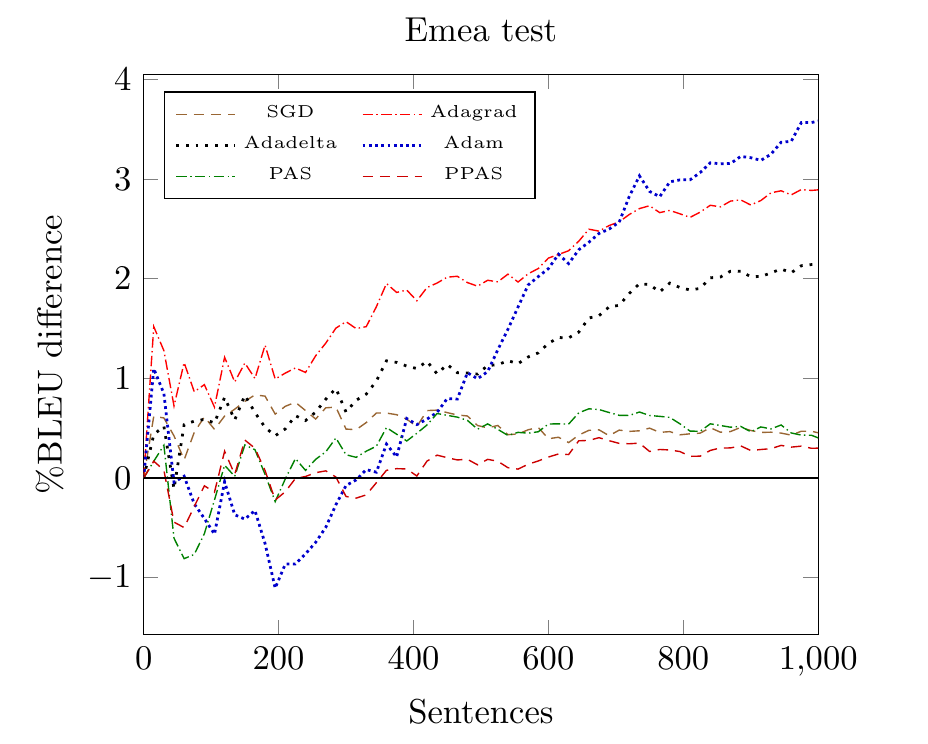}
	\caption{\label{fig:evol-bleu1} Evolution of a system trained on Europarl with the Emea test set. We show the BLEU difference (averaged up to the $n$-th sentence) with respect to an offline system.}
\end{figure}

First, we study the evolution of the different learning algorithms. \cref{fig:evol-bleu1} shows the difference in terms of BLEU between the baseline and the online systems, for the Emea test set as a function of the number of processed sentences. We show the average of BLEU up to the $n$-th sentence. 

As we can see in~\cref{fig:evol-bleu1}, the vanilla SGD algorithm remained steady along the learning process. Adaptive algorithms (Adagrad, Adadelta, Adam) had a more unstable behavior at the early learning stages. This is because these algorithms rely on the accumulation of the past gradients for updating the model weights. Therefore, at the beginning, they present a chaotic behavior, but once they are stabilized, the learning improves. The Adam algorithm clearly exhibits this: it took 400 sentences to improve the baseline, but from here, the enhancement was acute, eventually yielding the best overall performance. On the other hand, PA algorithms remain close to the original system, which denotes an excessive passivity. 

\begin{figure}
	\centering
	\includegraphics[width=0.5\textwidth]{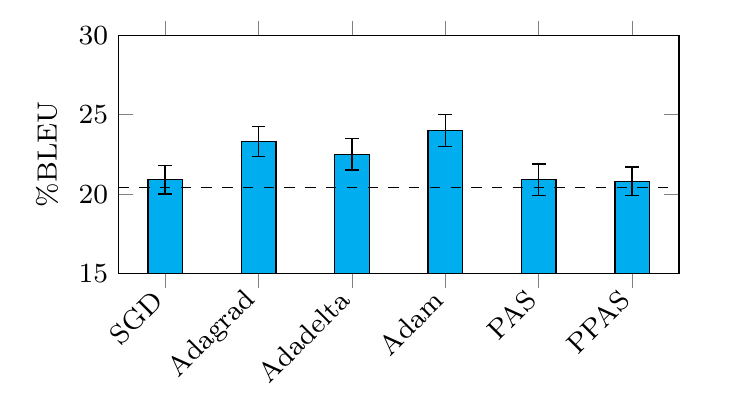}
	\caption{\label{fig:emea-europarl} Effect of tuning an NMT system trained on Europarl with the Emea test set in terms of BLEU. The horizontal line refers to an offline system. We show $95\%$ confidence intervals.}
\end{figure}

\cref{fig:emea-europarl} shows the overall differences between OL algorithms in terms of BLEU. The dashed line represents our baseline. All three adaptive SGD algorithms were able to significantly improve over the baseline. Vanilla SGD also yielded small enhancements, although they were non-significant. PA-like algorithms obtained a similar performance to the baseline.

Finally, \cref{tab:no-in-domain-nmt} condenses the results of the adaptation without an in-domain corpus. For the sake of clarity, we only show results for the best performing algorithm for each task. In almost every case, the online system performed significantly better than the offline one. The Emea and TED tasks had a similar difficulty for the out-of-domain system and the improvements obtained through OL were alike.

The NMT system performed much worse in the XRCE task. This is probably due to the corpus characteristics: since it is extracted from printer manuals, many of its structures are nonexistent in the out-of-domain corpus. Therefore, the performance drops. OL also succeeded in adapting the model to this task, but its contribution was limited by the size of the test set: OL could adapt the model to the test set, but only to some extent.

\noindent\begin{table}[h]
	\caption{\label{tab:no-in-domain-nmt} Offline vs online systems, trained without in-domain data. Bold results indicate a significant improvement over the offline system. We also show the best performing algorithm for each task.}
	\centering
	\footnotesize
	{\setlength{\tabcolsep}{4pt}
		\begin{tabular}{lllll}
			\toprule
			& &  Emea & TED & XRCE \\
			\cmidrule(lr){3-5}
			\multirow{3}{*}{\rotatebox[origin=c]{90}{\parbox[c]{1cm}{\centering Offline}}} & BLEU & $20.5 \pm 1.0$ & $20.9\pm 0.8$ & $8.3 \pm 0.9$\\
			& Meteor & $42.4 \pm 1.0$ & $41.1 \pm 0.9$ & $19.7 \pm 1.0$\\
			& TER & $69.0 \pm 1.6$  & $67.8 \pm 1.2$ & $103.8 \pm 3.5$ \\
			\midrule
			\multirow{4}{*}{\rotatebox[origin=c]{90}{\parbox[c]{1cm}{\centering Online}}} & BLEU & $\mathbf{24.0 \pm 1.0}$ & $\mathbf{24.0 \pm 0.9}$ & $ \mathbf{11.8 \pm 1.2}$\\
			& Meteor & $\mathbf{46.5 \pm 1.0}$ & $\mathbf{43.0 \pm 0.9}$& $19.8 \pm 1.3$\\
			& TER & $\mathbf{63.1 \pm 1.5}$ & $\mathbf{62.4 \pm 1.1}$ &$\mathbf{83.2 \pm 3.9}$ \\
			\cmidrule(lr){2-5}
			& Algorithm & Adam & Adadelta& Adam \\
			\bottomrule
		\end{tabular}}
	\end{table}

\subsection{Adapting with an in-domain}

In this case, we assume that we have a collection of in-domain data, in addition to our out-of-domain corpus. Therefore, we first train a general system on the out-of-domain data and fine-tune it with the in-domain training set. Next, we study if OL can still improve the model, although it is already adapted to the in-domain task. As before, we apply OL techniques to the test set.

\begin{figure}[!ht]
	\centering
	\includegraphics[width=0.5\textwidth]{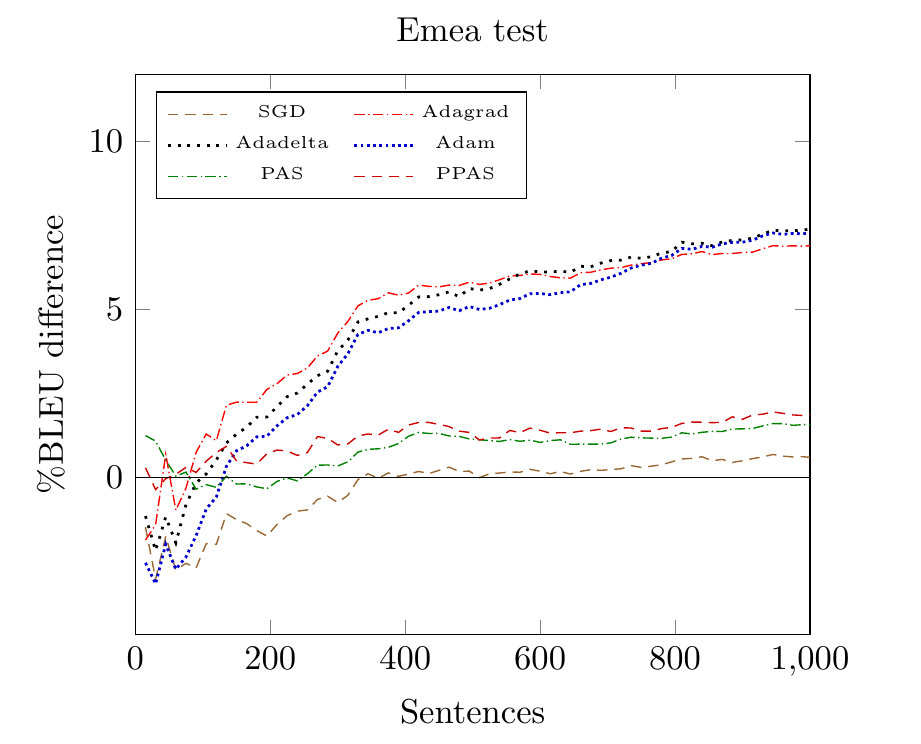}
	\caption{\label{fig:evol-bleu_europarl_emea_emea} Evolution of a system trained on Europarl and Emea with the Emea test set. We show the BLEU difference (averaged up to the $n$-th sentence) with respect to an offline system.}
\end{figure}

\cref{fig:evol-bleu_europarl_emea_emea} shows the BLEU differences between OL algorithms. In this case, the behavior was more stable for all algorithms. Although some variance can still be observed at the early stages of the process, the system learned faster and better than in the previous case. Since the effective learning started early, the cumulated enhancements were larger. From~\cref{fig:evol-bleu_europarl_emea_emea} we can also extract the importance of the adaptive SGD implementations, which make the learning process to converge faster and to a better point.

\begin{figure}[!h]
	\centering
	\includegraphics[width=0.5\textwidth]{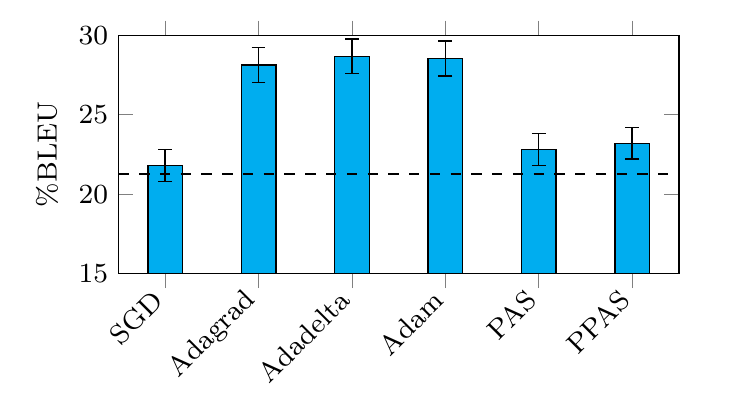}
	\caption{\label{fig:europarl_emea_emea} Effect of OL in an NMT system trained on Europarl and Emea. We use the Emea test set for applying OL. We show BLEU and $95\%$ confidence intervals. The horizontal line refers to the offline system baseline.}
\end{figure}

Absolute BLEU improvement is shown in~\cref{fig:europarl_emea_emea}. Again, the dashed line represents the offline baseline. In this case, the gains obtained by the adaptive SGD algorithms are even higher than in the previous experiment. Adagrad, Adadelta and Adam yielded to almost the same value, producing improvements of more than 7 BLEU points with respect to the baseline.

\cref{tab:in-domain-nmt} shows the overall results for this scenario. Again, we only show figures of the best method for each task.  The use of in-domain training data allowed the NMT system to yield much better translation quality than in the previous case. Especially dramatics were the improvements in the TED and XRCE corpora (+10.4 and +26.4 BLEU points, respectively). For such tasks, the use of OL only affected the system to a negligible extent.

On the other hand, the improvements observed in the Emea task were smaller than those obtained in other corpora. Nevertheless, this is clearly complemented by the online system. In this case, OL was able to adapt the system to the test set. The system was significantly improved. The post-editing effort was reduced approximately by 15\%.

\noindent\begin{table} [h]
	\caption{\label{tab:in-domain-nmt}Offline vs online systems, trained with out-of-domain data and adapted with additional in-domain data. Bold results indicate a significant improvement of the online system with respect to the offline. We also indicate the best performing algorithm for each task.}
	\centering
	\footnotesize
	{\setlength{\tabcolsep}{4pt}
	\begin{tabular}{lllll}
		\toprule
		& &  Emea & TED & XRCE \\
		\cmidrule(lr){3-5}
		\multirow{3}{*}{\rotatebox[origin=c]{90}{\parbox[c]{1cm}{\centering Offline}}}& BLEU & $21.3\pm 1.0$ & $31.3\pm 1.0$ & $ 34.7 \pm 2.2$\\
		& Meteor & $39.5 \pm 1.1$ & $52.0 \pm 1.0$ & $52.2 \pm 2.3$\\
		& TER & $66.9 \pm 1.5$  & $52.3 \pm 1.0$ & $55.0 \pm 2.4$ \\
		\midrule
		\multirow{4}{*}{\rotatebox[origin=c]{90}{\parbox[c]{1cm}{\centering Online}}}  & BLEU & $\mathbf{28.6 \pm 1.2}$ &  $31.4 \pm 1.0$& $35.1 \pm 2.2$\\
		& Meteor & $\mathbf{48.8 \pm 1.1}$ & $51.8 \pm 0.9$ & $53.0\pm 2.1$\\
		& TER & $\mathbf{57.0 \pm 1.5}$ & $52.2  \pm 1.0$ & $55.8 \pm 2.2$\\
		\cmidrule(lr){2-5}
		& Algorithm & Adam & PPAS & PPAS \\
		\bottomrule
	\end{tabular}}
\end{table}

\subsection{Refining an NMT system}

Finally, we move to the scenario in which we have enough in-domain data for training an NMT system. The system is trained exclusively on the in-domain data. OL learning techniques can be used for refining the system at test time.

\begin{figure}[!ht]
	\centering
	\includegraphics[width=0.5\textwidth]{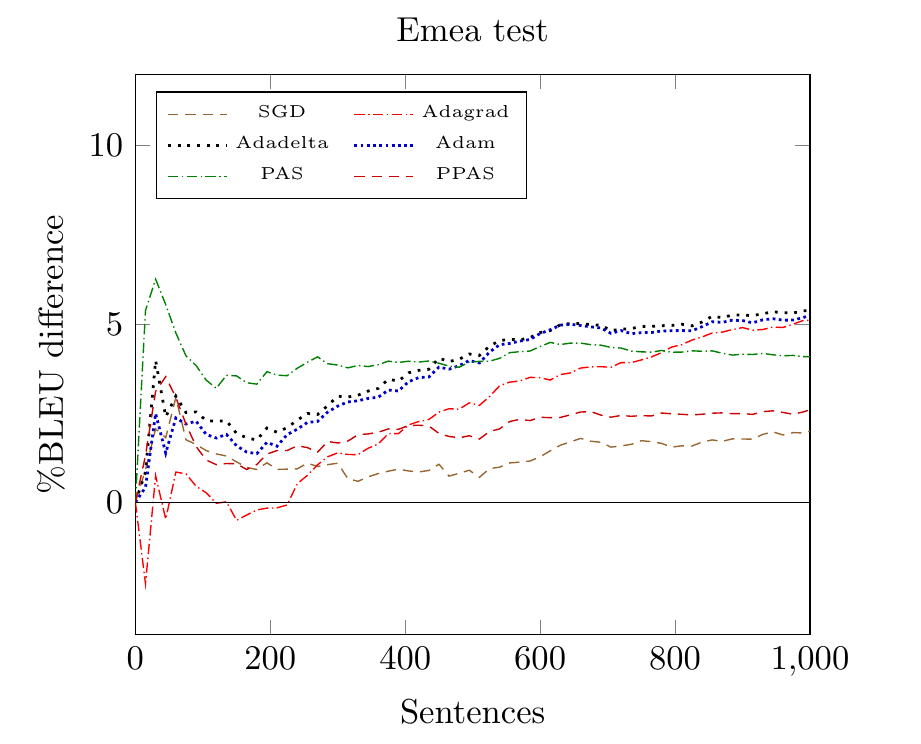}
	\caption{\label{fig:evol-bleu_emea} Evolution of a system trained on Europarl and Emea with the own Emea test set. We show the BLEU difference (averaged up to the $n$-th sentence) with respect to an offline system.}
\end{figure}

\cref{fig:evol-bleu_emea} shows the evolution of the BLEU scores along the online learning process. Again, algorithms with adaptive learning rate perform better than the rest of them. Nevertheless, such differences are lower than in other tasks. The PAS algorithm is especially effective at the first iterations of the process. %TODO: Why?

\begin{figure}[h]
	\centering
	\includegraphics[width=0.5\textwidth]{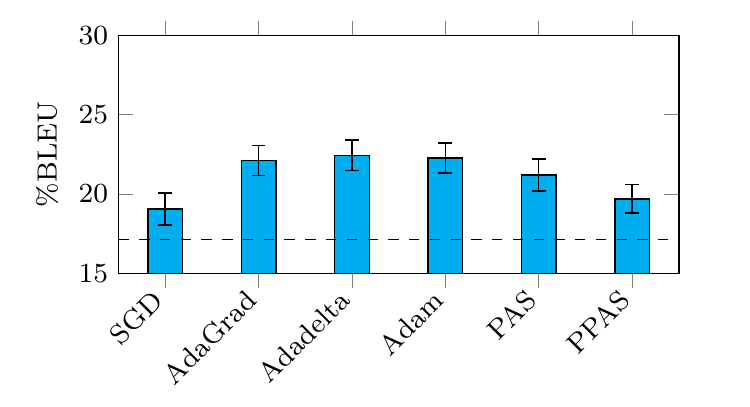}
	\caption{\label{fig:emea} Effect of tuning an NMT system trained on Emea and refined with the own Emea test, in terms of BLEU. The horizontal line refers to an offline system. We show $95\%$ confidence intervals.}
\end{figure}

All OL algorithms significantly outperform the offline baseline, as shown in~\cref{fig:emea}. The performance of the PAS algorithm is comparable to the best-performing algorithms. The projected PA version is slightly better than vanilla SGD, although it yields worse results than the other algorithms.

 \cref{tab:only-in-domain} collapses the performance of the systems in the different tasks.

\begin{table} [h]
	\caption{\label{tab:only-in-domain} Offline vs online systems, trained solely on the in-domain data. Bold results indicate a significant improvement of the online system with respect to the offline. We also indicate the best performing algorithm for each task.}
	\centering
	\footnotesize
	{\setlength{\tabcolsep}{4pt}
	\begin{tabular}{lllll}
		\toprule
		& &  Emea & TED & XRCE \\
		\cmidrule(lr){3-5}
		\multirow{3}{*}{\rotatebox[origin=c]{90}{\parbox[c]{1cm}{\centering Offline}}} & BLEU & $17.6 \pm 0.9$  & $27.1\pm 0.9$ & $31.0\pm 2.2$\\
		& Meteor & $37.1 \pm 0.9$ & $47.6 \pm 1.0$ & $49.8 \pm 2.0$\\
		& TER & $74.2 \pm 2.2$  & $57.9 \pm 1.0$& $62.8 \pm 2.7$\\
		\midrule
		\multirow{4}{*}{\rotatebox[origin=c]{90}{\parbox[c]{1cm}{\centering Online}}}  & BLEU & $\mathbf{22.4 \pm 1.0}$ & $28.2 \pm 0.9$ & $\mathbf{36.7 \pm 2.3}$\\
		 & Meteor & $\mathbf{41.1 \pm 0.9}$ & $49.1 \pm 1.0$ & $\mathbf{54.2 \pm 2.0}$ \\
		& TER & $\mathbf{63.8 \pm 1.4}$ & $56.0\pm 1.1$ & $\mathbf{56.5 \pm 2.6}$\\
		\cmidrule(lr){2-5}
		& Algorithm & Adadelta & Adadelta & Adagrad \\
		\bottomrule
	\end{tabular}}
\end{table}

Even when the systems have been exclusively trained on in-domain data, OL helps to perform a more fine-grained adaptation to the text to translate. In the Emea and XRCE tasks we could observe significant improvements (approximately 5 points of BLEU). This is probably due to regularities in the test set, which allow to exploit OL to its full. OL also improved the performance in TED task, although to a lower extent.

\section{Conclusions and future work}
 \label{sec:conclusions}

In this work, we studied the implementation of OL strategies for NMT. We empirically demonstrated the capacity of NMT for adapting to new domains, by means of gradient descent. Moreover, we discussed and compared a wide range of gradient descent algorithms. In addition, we proposed two novel methods, inspired by a PA strategy and tackled by means of subgradient optimization methods. PA-based methods offer a competitive performance. In almost every case, they outperformed vanilla SGD. Nevertheless, adaptive SGD algorithms performed generally better than PA. We found two cases in which the performance of the PA method was superior to adaptive SGD algorithms, although differences were small. 

One of the main drawbacks relating the NMT is the independence (up to some point) of the network objective function (probability of the target sentence given the source) with respect to the evaluation metric (e.g. BLEU or TER). Minimum risk training  aims to overcome this gap and it has successfully been applied to classical SMT~\citep{Och03,Chiang12} and recently to NMT~\citep{Shen16}. 

In this work, we set the basis for using BLEU or other non-differentiable loss function, since our PAS and PPAS algorithms cope with non-differentiable functions. As future work, we intend to directly optimize the evaluation metric, integrating it into the online learning framework~\citep{Martinez-Gomez12}. A major challenge is keeping the update time at a reasonable time.

Finally, interactive machine translation (IMT) is tightly related with this work. IMT consists in an evolution of the classical post-editing scenario, striving for profiting the human effort made in the post-editing process. Under the IMT paradigm, human and computer collaborate in order to minimize  the user effort. The development of interactive NMT systems has been recently addressed~\citep{Knowles16,Peris16}. The inclusion of OL techniques into the interactive framework is the next natural step to take, in order to develop more adaptive and productive translation systems.

\section*{Acknowledgments}

The research leading to these results has received funding from
the Generalitat Valenciana under grant PROMETEOII/2014/030. We also acknowledge NVIDIA for the donation of a GPU used in this work.

\bibliography{emnlp2017}
\bibliographystyle{emnlp_natbib}

\end{document}